\documentclass[10pt,twocolumn,letterpaper]{article}

\usepackage{wacv}
\usepackage{times}
\usepackage{epsfig}
\usepackage{graphicx}
\usepackage{amsmath}
\usepackage{amssymb}
\usepackage{booktabs}
\usepackage{threeparttable}
\usepackage{footnote}
\makesavenoteenv{tabular}
\makesavenoteenv{table}
\usepackage{colortbl}
\usepackage{multirow}

\def\wacvPaperID{1021} 

\wacvfinalcopy 

\ifwacvfinal
\def\assignedStartPage{1} 
\fi

\ifwacvfinal
\usepackage[breaklinks=true,bookmarks=false,colorlinks]{hyperref}
\else
\usepackage[pagebackref=true,breaklinks=true,colorlinks,bookmarks=false]{hyperref}
\fi

\usepackage[dvipsnames]{xcolor}
\definecolor{OliveGreen}{rgb}{0,0.6,0}
\definecolor{SoftRed}{rgb}{1,0.2,0.2}
 \hypersetup{
      linkcolor=Magenta,
      filecolor=blue,
      citecolor = OliveGreen,      
      urlcolor=Magenta
      }

\ifwacvfinal
\else
\fi

\begin{document}
\definecolor{SkyBlue}{rgb}{0.88,1,1}
\title{Hear Me Out: Fusional Approaches for Audio Augmented Temporal Action Localization}

\author{Anurag Bagchi\\
CVIT, IIIT Hyderabad\\
{\tt\small miccooper9@gmail.com}
\and
Jazib Mahmood\\
CVIT, IIIT Hyderabad\\
{\tt\small jazib.mahmood@research.iiit.ac.in}
\and
Dolton Fernandes\\
CVIT, IIIT Hyderabad\\
{\tt\small dolton.fernandes@research.iiit.ac.in}
\and
Ravi Kiran Sarvadevabhatla\\
CVIT, IIIT Hyderabad\\
{\tt\small ravi.kiran@iiit.ac.in}
}

\maketitle

\begin{abstract}
   State of the art architectures for untrimmed video Temporal Action Localization (TAL) have only considered RGB and Flow modalities, leaving the information-rich audio modality totally unexploited. Audio fusion has been explored for the related but arguably easier problem of trimmed (clip-level) action recognition. However, TAL poses a unique set of challenges. In this paper, we propose simple but effective fusion-based approaches for TAL. To the best of our knowledge, our work is the first to jointly consider audio and video modalities for supervised TAL. We experimentally show that our schemes consistently improve performance for state of the art video-only TAL approaches. Specifically, they help achieve new state of the art performance on large-scale benchmark datasets - ActivityNet-1.3 (54.34 mAP@0.5) and THUMOS14 (57.18 mAP@0.5). Our experiments include ablations involving multiple fusion schemes, modality combinations and TAL architectures. Our code, models and associated data are available at \url{https://github.com/skelemoa/tal-hmo}. 
\end{abstract}

\section{\uppercase{Introduction}}
\label{sec:introduction}

With the boom in online video production, video understanding has become one of the most heavily researched domains. Temporal Action Localization (TAL) is one of the most interesting and challenging problems in the domain. The objective of TAL is to identify the category (class label) of activities present in a long, untrimmed, real world video and their temporal boundaries (start and end time). Apart from inheriting the challenges from the related problem of \textit{trimmed} (clip-level) video action recognition, TAL also requires accurate temporal segmentation, i.e. to precisely locate the start time and end time of action categories present in a given video.

TAL is an active area of research and several approaches have been proposed to tackle the problem~\cite{Xu_2020_CVPR,PGCN2019ICCV,DBLP:journals/corr/abs-2011-11479,Liu_2021_CVPR,Liu_2021_CVPR,Lin_2021_CVPR}. For the most part, existing approaches depend solely on the visual modality (RGB, Optical Flow). An important and obvious source of additional information  -- the audio modality -- has been overlooked. This is surprising since audio has been shown to be immensely useful for other video-based tasks such as object localization~\cite{8237335}, action recognition~\cite{DBLP:conf/eccv/OwensE18,10.1145/2964284.2964328,8578915,b1bace6f29a746caa54afb1f42bcbf36,DBLP:journals/ijcv/OwensWMFT18} and egocentric action recognition~\cite{9010900}. 

\begin{figure*}
\centering
\includegraphics[width=\textwidth]{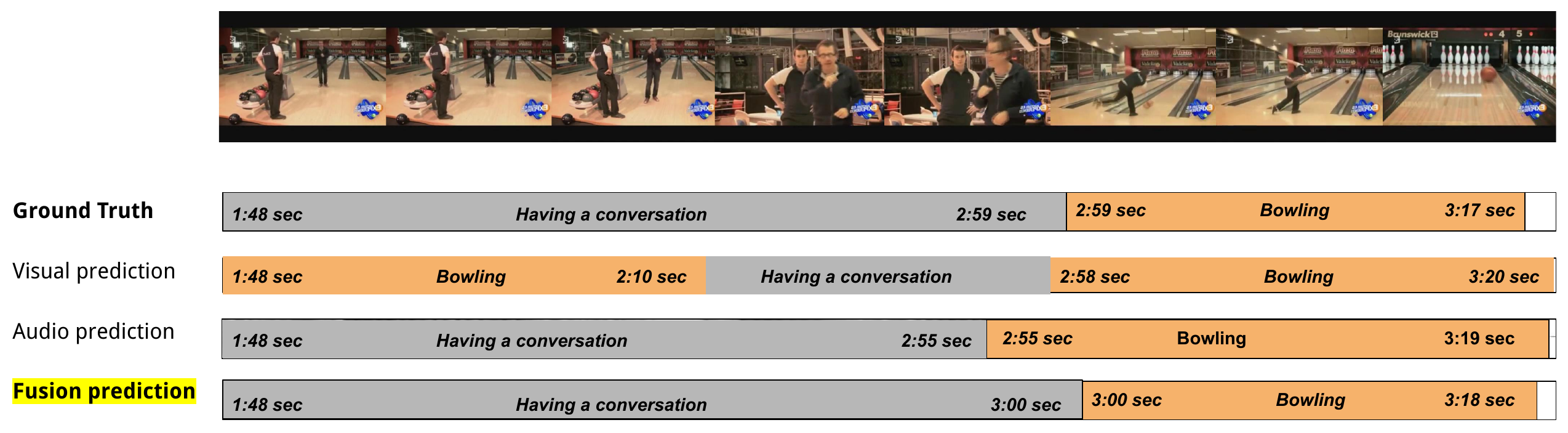}
\caption{An example illustrating a scenario where audio modality can help improve peformance over video-only temporal action localization.} 
\label{fig:audio-helps-tal}
\end{figure*}

Analyzing the untrimmed videos, it is evident that the audio track provides crucial complementary information regarding the action classes and their temporal extents. Action class segments in untrimmed videos are often characterized by signature audio transitions as the activity progresses (e.g. the rolling of a ball in a bowling alley culminating in striking of pins, an aquatic diving event culminating with the sound of splash in the water). Depending on the activity, the associated audio features can supplement and complement their video counterparts if the two feature sequences are fused judiciously (Figure~\ref{fig:audio-helps-tal}).

Motivated by observations mentioned above, we make the following contributions:
\begin{itemize}
    \item We propose simple but effective fusion approaches to combine audio and video modalities for TAL (Section~\ref{sec:fusionschemes}). Our work is the first to jointly process audio and video modalities for supervised TAL.
    \item We show that our fusion schemes can be readily plugged into existing state-of-the-art video based TAL pipelines (Section~\ref{sec:fusionschemes}).
    \item To determine the efficacy of our fusional approaches, we perform comparative evaluation on large-scale benchmark datasets - ActivityNet and THUMOS14. Our results (Section~\ref{sec:results}) show that the proposed schemes consistently boost performance for state of the art TAL approaches, resulting in an improved mAP of 52.73 for ActivityNet-1.3 and 57.18 mAP for THUMOS14.
    \item Our experiments include ablations involving multiple fusion schemes, modality combinations and TAL architectures. 
\end{itemize}

Our code, models and additional data can be found at \url{https://github.com/skelemoa/tal-hmo}.

\section{\uppercase{Related Work}}
\label{sec:related}

\noindent \textbf{Temporal Action Localization:} A popular technique for Temporal Action Localization is inspired from the proposal-based approach for object detection~\cite{girshick2015fast}. In this approach, a set of so-called proposals are generated and subsequently refined to produce the final class and boundary predictions. Many recent approaches employ this proposal-based formulation~\cite{DBLP:journals/corr/ShouWC16,DBLP:journals/corr/ZhaoXWWLT17,DBLP:journals/corr/XuDS17}. Specifically, this is the case for  state-of-the-art approaches we consider in this paper -- G-TAD~\cite{Xu_2020_CVPR}, PGCN~\cite{PGCN2019ICCV} and MUSES baseline~\cite{Liu_2021_CVPR}. Both G-TAD~\cite{Xu_2020_CVPR} and PGCN~\cite{PGCN2019ICCV} use graph convolutional networks and the concept of edges to share context and background information between proposals. MUSES baseline~\cite{Liu_2021_CVPR} on the other hand, achieves state of the art results on the benchmark datasets by employing a temporal aggregation module, originally intended to account for the frequent camera changes in their new multi-shot dataset. 

\begin{figure*}[!t]
    \centering
    \noindent
    \includegraphics[width=\linewidth]{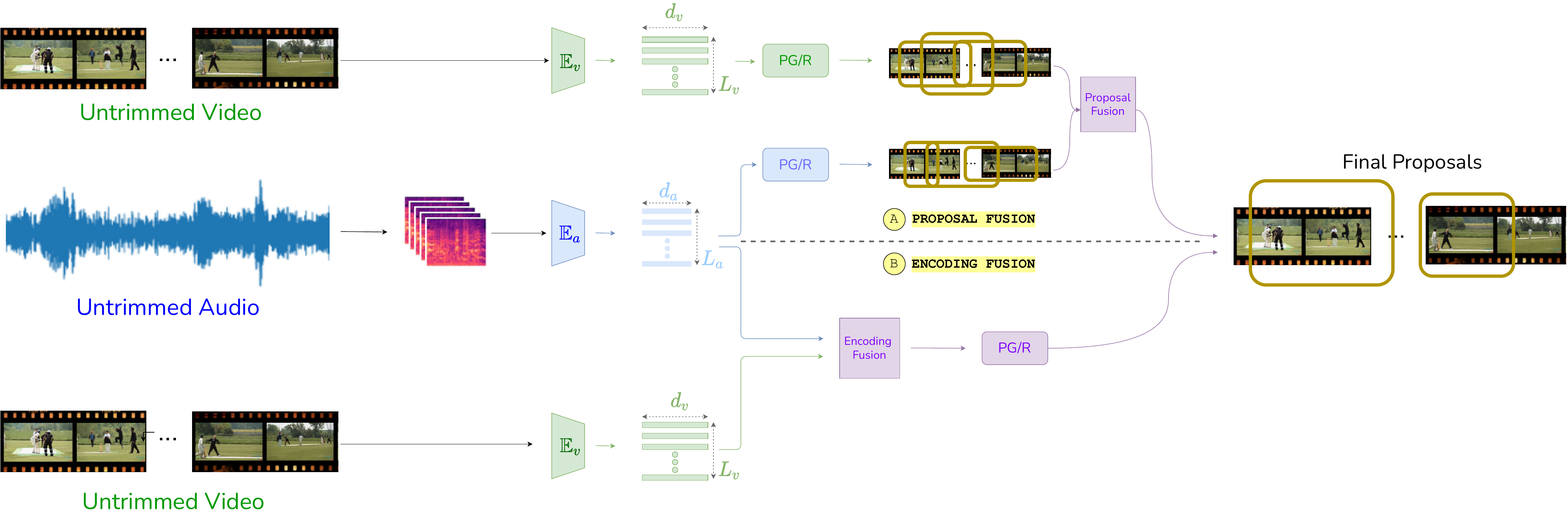}
    \caption{An illustrative overview of our fusion schemes (Section~\ref{sec:fusionschemes}).}
    \label{fig:avfusion}
\end{figure*}

The proposal generation schemes in the literature are either anchor-based~\cite{DBLP:journals/corr/GaoYSCN17,DBLP:journals/corr/abs-1807-04821,DBLP:journals/corr/abs-1811-11524} or generate a boundary probabilty sequence~\cite{DBLP:journals/corr/abs-1806-02964,DBLP:journals/corr/abs-1907-09702,DBLP:journals/corr/abs-2009-07641}. Past work in this domain also includes end to end techniques~\cite{689ad8d461864d1b9f89ea6ecee65fe2,DBLP:journals/corr/YeungRMF15,DBLP:journals/corr/abs-1710-06236} which combine the two stages. Frame-level techniques which require merging steps to generate boundary predictions also exist~\cite{DBLP:journals/corr/ShouCZMC17,DBLP:journals/corr/MontesSN16,8737877}. We augment the proposal-based state of the art approaches designed solely for visual modality by incorporating audio into their architectures.

\noindent \textbf{Audio-only based Localization:} Speaker diarization~ \cite{8462628,8683892} involves localization of speaker boundaries and grouping segments that belong to the same speaker. The DCASE Challenge~\cite{DCASE2020Workshop} examines sound event detection in domestic environments as one of the challenge tasks~\cite{Miyazaki2020,Hao2020,Ebbers2020,Lin2019,Delphin-Poulat2019,Shi2019}. In our action localization setting, note that audio modality is unrestricted. It is not confined to speech or labelled sound events which is the case for audio-only localization.

\noindent \textbf{Audio-visual localization:} This task is essentially localization of specific events of interest across modalities. Given a temporal segment of one modality (auditory or visual), we would like to localize the temporal segment
of the associated content in the other modality. This is different from our task of temporal action localization (TAL) which focuses on predicting the class-labels and temporal segment boundaries of all actions present in a given video.

\noindent \textbf{Fusion approaches for TAL:} Fusion of multiple modalities is an effective technique for video understanding tasks due to its ability to incorporate all the information available in videos. The fusion schemes present in the literature can be divided into 3 broad categories -- \textit{early fusion}, \textit{mid fusion} and \textit{late fusion}. 

\textit{Late fusion} combines the representations closer to the output end from each individual modality stream. When per-modality predictions are fused, this technique is also referred to as \textit{decision level fusion}. \textit{Decision level fusion} is used in I3D~\cite{8099985} which serves as a feature extractor for the current state-of-the-art in TAL. However, unlike the popular classification setting, \textit{decision level fusion} is challenging for TAL since predictions often differ in relative temporal extents. PGCN~\cite{PGCN2019ICCV}, introduced earlier, solves this problem by performing Non Maximal Suppression on the combined pool of proposals from the two modalities (RGB, Optical Flow). MUSES baseline~\cite{Liu_2021_CVPR} fuses the RGB and Flow predictions. \textit{Mid fusion} combines mid-level feature representations from each individual modality stream. Feichtenhofer et. al.~\cite{feichtenhofer2016convolutional} found that fusing RGB and Optical Flow streams at the last convolutional layer yields good visual modality features. The resulting mid-level features have been successfully employed by well performing TAL approaches~\cite{Lin_Li_Wang_Tai_Luo_Cui_Wang_Li_Huang_Ji_2020,DBLP:journals/corr/abs-1907-09702,DBLP:journals/corr/abs-1806-02964,li2019deep}. In particular, they are utilized by G-TAD~\cite{PGCN2019ICCV} to obtain feature representations for each temporal proposal. \textit{Early fusion} involves fusing the modalities at the input level. In the few papers that compare different fusion schemes~\cite{inbook,tian2018audiovisual}, \textit{early fusion} has been shown to be generally an inferior choice. 

Apart from within (visual) modality fusion mentioned above, audio-visual fusion specifically has been shown to benefit (trimmed) clip-level action recognition~\cite{10.1145/2964284.2964328,8578915,b1bace6f29a746caa54afb1f42bcbf36,9010900} and audio-visual event localization~\cite{zhou2021positive,he2021multimodal,xu2020cross}. The audio modality has also been shown to be beneficial for the weakly supervised version of TAL~\cite{lee2021crossattentional} wherein the boundary labels for activity instances are absent. However, the lack of labels is a fundamental performance bottleneck compared to the supervised approach. 

In our work, we introduce two \textit{mid-level fusion} schemes along with \textit{decision level fusion} to combine Audio, RGB, Flow modalities for state of the art supervised TAL.  

\section{\uppercase{Proposed Fusion Schemes}}
\label{sec:fusionschemes}

\noindent \textbf{Processing stages in video-only {TAL}:} Temporal Action Localization can be formulated as the task of predicting start and end times $(t_{s},t_{e})$ and action label $a$ for each action in an untrimmed RGB video $V \in \mathbb{R}^{F \times 3 \times H \times W}$, where $F$ is the number of frames, $H$ and $W$ represent the frame height and width. Despite the architectural differences, state of the art TAL approaches typically consist of three stages: \textit{feature extraction}, \textit{proposal generation} and \textit{proposal refinement}.

The \textit{feature extraction} stage transforms a video into a sequence of feature vectors corresponding to each visual modality (RGB and Flow). Specifically, the feature extractor operates on fixed size snippets $S \in \mathbb{R} ^{L \times C \times H \times W}$ and produces a feature vector $f \in \mathbb{R}^{d_v}$. Here, $C$ is the number of channels and $L$ is the number of frames in the snippet. This results in the feature vector sequence $F_v \in \mathbb{R}^{L_{v} \times d_v}$ mentioned above where $L_v$ is the number of snippets. This stage is shown as the component shaded green (containing $\mathbb{E}_v$) in Figure~\ref{fig:avfusion}.

The \textit{proposal generation} stage processes the feature sequence mentioned above to generate action proposals. Each candidate action proposal is associated with temporal extent (start and end time) relative to the input video, and a confidence score. In some approaches, each proposal is also associated with an activity class label.

The \textit{proposal refinement} stage takes the feature sequence corresponding to each candidate proposal as input and refines the boundary predictions and confidence scores. Some approaches modify the class label predictions as well. The final proposals are generally obtained by applying non maximal suppression to weed out the redundancies arising from highly overlapping proposals. Also, note that some approaches do not treat proposal generation and refinement as two different stages. To accommodate this variety, we depict the processing associated with proposal generation and refinement as a single module titled `PG/R' in Figure~\ref{fig:avfusion}.

\subsection{Incorporating Audio}

 As with video-only TAL approaches, the first stage of audio modality processing consists of extracting a sequence of features from audio snippets (refer to the blue shaded module termed $\mathbb{E}_a$ in Figure~\ref{fig:avfusion}). This results in the `audio' feature vector sequence $F_a \in \mathbb{R}^{L_{a} \times d_a}$ mentioned above where $L_a$ is the number of audio snippets and $d_a$ is the length of feature vector for an audio snippet. 
 
 Our objective is to incorporate audio as seamlessly as possible into existing video-only TAL architectures. To enable flexible integration, we present two schemes - \textit{proposal fusion} and \textit{encoding fusion}. We describe these schemes next.

\subsection{ \textcircled{A} Proposal fusion} 
\label{sec:propfusion}

This is a \textit{decision fusion} approach and as suggested by its name, the basic idea is to merge proposals from the audio and video modalities (see `Proposal Fusion' in Figure~\ref{fig:avfusion}). To begin with, audio proposals are obtained in a manner similar to the procedure used to obtain video proposals. As mentioned earlier, it is straightforward to fuse action class predictions from each modality by simply averaging the probabilities. However, TAL proposals consist of regression scores for action boundaries, where averaging across modality does not make much sense since it is likely to introduce error into the predictions. This makes the fusion task challenging in the TAL setting. 

To solve this problem while adhering to our objective of leaving the existing video-only pipeline untouched, we re purpose the corresponding module from the pipelines. Specifically, we use Non-Maximal Suppression (NMS) for iteratively choosing the best proposals which minimally overlap with other proposals. In some architectures (e.g. PGCN~\cite{PGCN2019ICCV}), NMS is applied to separate proposals from RGB and Flow components which contribute together as part of the visual modality. We extend this, by initially pooling visual modality proposals with audio proposals, and then apply NMS.

\subsection{ \raisebox{.5pt}{\textcircled{\raisebox{-1.01pt} {B}}} Encoding fusion}
\label{sec:encfusion}

Instead of the late fusion of per-modality proposals described above, an alternative is to utilize the combined set of audio $F_a$ and video feature sequences $F_v$ to generate a single, unified set of proposals. However, since the encoded representation dimensions $d_a,d_v$ and the number of sequence elements $L_a,L_v$ can be unequal, standard dimension-wise concatenation techniques are not applicable. To tackle this issue, we explore four approaches to make the sequence lengths equal for feature fusion (depicted as purple block titled `Encoding Fusion' in Figure~\ref{fig:avfusion}). 

For the first two approaches, we revisit the feature extraction phase and extract audio features at the frame rate used for videos. As a result, we obtain a paired sequence of audio and video snippets (i.e. $L_a = L_v$).

\begin{figure}[!t]
    \centering
    \noindent
    \includegraphics[width=\linewidth]{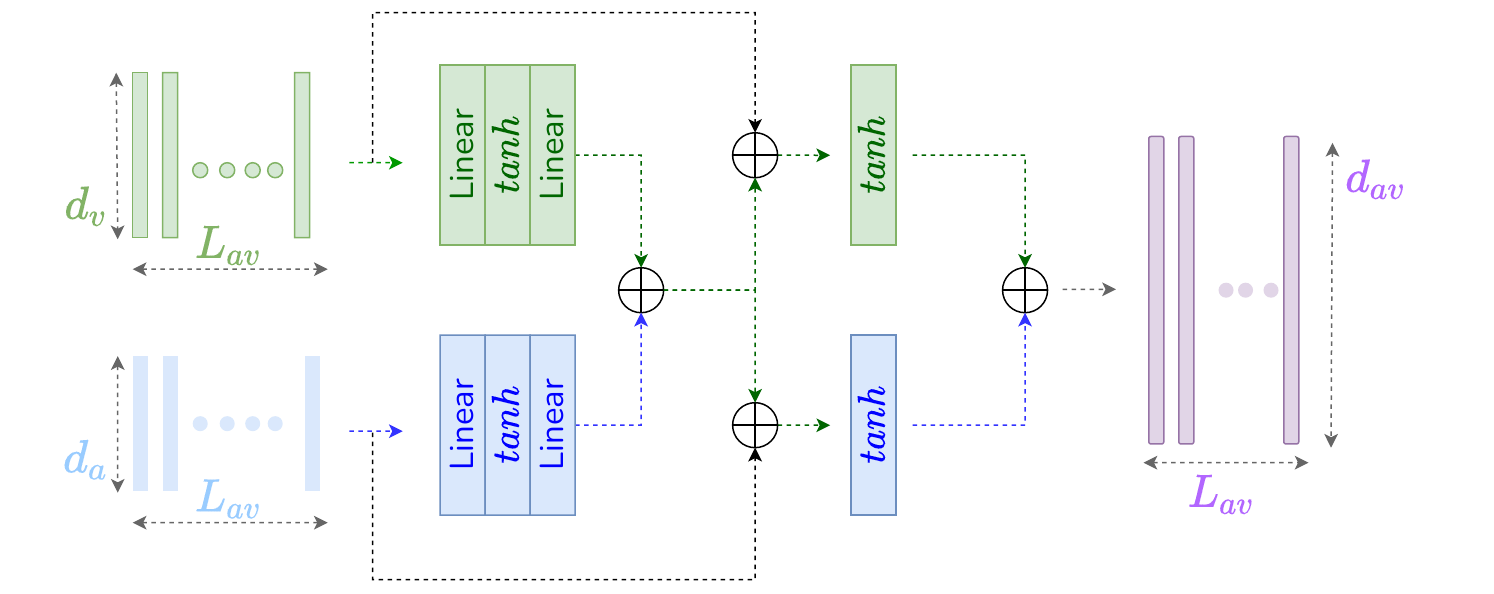}
    \caption{Residual Multimodal Attention mechanism with video-only and audio features as a form of encoding fusion (Section~\ref{sec:encfusion}).  $\bigoplus$ indicates tensor addition.}
    \label{fig:mmatt}
\end{figure}

\begin{itemize}
\item \textbf{Concatenation (Concat):} The paired sequences are concatenated along the feature dimension.

\item \textbf{Residual Multimodal Attention (RMAttn):}  To refine each modality's representation using features from other modality, we employ a residual multimodal attention mechanism~\cite{tian2018audiovisual} as shown in Figure~\ref{fig:mmatt}. 

We define the attention function $f_{att}$ and it can be adaptively learned from the visual feature map $v_t$ and audio feature vector $a_t$. At each time step t, the visual context vector $v_t^{att}$is computed by:

\begin{equation}
v_t^{att} = f_{att}(a_t, v_t) = \sum_{i=1}^{k} w_t^it_t^i ,
\end{equation}
where $w_t$ is an attention weight vector corresponding to the probability distribution over $k$ visual regions that are attended by its audio counterpart. The
attention weights can be computed based on MLP with a Softmax activation
function:
\begin{equation}
    w_t = Softmax(x_t) ,
\end{equation}
\begin{equation}    
x_t = W_f\sigma(W_vU_v(v_t) + (W_aU_a(a_t)){1}^T) ,
\end{equation}

where $U_v$ and $U_a$, implemented by a dense layer with nonlinearity, are two transformation functions that project audio and visual features to the same dimension
$d, W_v \in R^{k\times d}, W_a \in R^{k\times d}, W_f \in R^{1\times k}$ are parameters, the entries in ${1}^T \in \mathbb{R}^k$ are all 1, $\sigma()$ is the hyperbolic tangent function, and $w_t \in \mathbb{R}^k$
is the computed attention map. 
\end{itemize}

The other two encoding fusion approaches:
\begin{itemize}
    
\item \textbf{Duplicate and Trim (DupTrim):} Suppose $L_v < L_a$ and $k=\frac{L_a}{L_v}$. We first duplicate each visual feature to obtain a sequence of length $k L_v$. We then trim both the audio and visual feature sequences to a common length $L_m = min(L_a,k L_v)$. A similar procedure is followed for the other case ($L_a < L_v$).  

\item \textbf{Average and Trim (AvgTrim):} Suppose $L_v < L_a$ and $k=\frac{L_a}{L_v}$. We group audio features into subsequences of length $k^{'} = \lceil k \rceil$. We then form a new feature sequence of length $L_a^{'}$ by averaging each $k^{'}$-sized group. Following a procedure similar to `DupTrim' above, we trim the modality sequences to a common length, i.e. $L_m = min(L_a^{'},L_v)$. 

\end{itemize}

For the above approaches involving trimming, the resulting audio and video sequences are concatenated along the feature dimension to obtain the fused multimodal feature sequence. 

For all the approaches, the resulting fused representation is processed by a `PG/R' (proposal generation and refinement) module to obtain the final predictions, similar to its usage mentioned earlier (see Figure~\ref{fig:avfusion}). Apart from the fusion schemes, we also note that each scheme involves additional choices. In our experiments, we perform comparative evaluation for all the resulting combinations.

\section{\uppercase{Implementation Details}}
\label{sec:Implement}

\noindent \textbf{TAL video-only architectures:} To determine the utility of audio modality and to ensure our method is easily applicable to any video-only TAL approach, we do not change the architectures and hyperparameters (e.g. snippet length, frame rate, optimizers) for the baselines. The feature extraction components of the baselines are summarized in Table~\ref{tab:archsetup}.

\begin{table*}[!t]
\centering
\begin{threeparttable}
\resizebox{\linewidth}{!}
{
\centering
\begin{tabular}{c|c|c|c|c|c|c|c}
       & \multicolumn{2}{c|}{ } & \multicolumn{2}{c|}{ } & \multicolumn{3}{c}{\textbf{+Audio}} \\
\toprule
Dataset & Setup Id & Architecture & Visual Features  & mAP & mAP & fusion scheme & fusion type \\
\toprule
\multirow{2}{*}[-0.75em]{\rule{0pt}{2ex} THUMOS14\cite{THUMOS14}}
&  1
&  GTAD\cite{Xu_2020_CVPR} & TSN\cite{8454294} & $40.20$ & $\mathbf{42.01}$ & encoding & Concat  \\
 & 2 & PGCN\cite{PGCN2019ICCV} & TSP~\cite{DBLP:journals/corr/abs-2011-11479} & $53.50$ & $\mathbf{53.96}$ & encoding & AvgTrim \\
 & 3 & MUSES\cite{Liu_2021_CVPR} & I3D\cite{8099985}  & $56.16$\tnote{1} & \cellcolor{SkyBlue}$\mathbf{57.18}$ & encoding & Concat \\
 \midrule
  \multirow{2}{*}[0em]{\rule{0pt}{2ex} ActivityNet-1.3\cite{Heilbron_2015_CVPR}}
  &  1
  & GTAD\cite{Xu_2020_CVPR} & GES\cite{xiong2016cuhk}  & $41.50$\tnote{2} & $\mathbf{42.17}$ & encoding & Concat \\
  &  2
  & GTAD\cite{Xu_2020_CVPR} & TSP~\cite{DBLP:journals/corr/abs-2011-11479} & $51.26$ & \cellcolor{SkyBlue}$\mathbf{54.34}$ & encoding & RMAttn \\
 \bottomrule
\end{tabular}
}

\caption{Architectural pipeline components for top-performing TAL approaches. To reduce clutter, only mAP@0.5 is reported. The `\textbf{+Audio}' group refers to the fusion configuration corresponding to the best results (Section~\ref{sec:fusionschemes}).}
\begin{tablenotes}\scriptsize
 \item[1] From our best run using the official code from \cite{Liu_2021_CVPR}. In the paper, reported mAP is $56.90$
 \item[2] The result is obtained by repeating the evaluation on the set of videos currently available.
\end{tablenotes}
\label{tab:archsetup}
\end{threeparttable}
\end{table*}

\noindent \textbf{Audio extraction:} For audio, we use VGGish~\cite{hershey2017cnn}, a state of the art approach for audio feature extraction. We use a sampling rate of 16kHz to extract the audio signal and extract 128-D features from 1.2s long snippets. For experiments involving attentional fusion and simple concatenation, we extract features by centering the 1.2s window about the snippets used for video feature extraction so as to maintain the same feature sequence length for audio and video modalities. Windows shorter than 1.2s (a few starting and ending ones) are padded with zeros at the end to specify that no more information is present and to match the 1.2s window requirement. Although the opposite (i.e. changing sampling rate for video, keeping the audio setup unchanged) is possible, we prefer the former since the video-only architecture and (video) data processing can be used as specified originally, without worrying about the consequences of such a change on the existing video-only architectural setup and hyperparameter choices.

\noindent \textbf{Proposal Generation and Refinement (PG/R):} For proposal generation, we consider state of the art architectures GTAD\cite{Xu_2020_CVPR}, BMN\cite{DBLP:journals/corr/abs-1907-09702} and BSN\cite{DBLP:journals/corr/abs-1806-02964}. Similarly, for proposal refinement, we consider proposals generated from BMN and BSN, refined in PGCN\cite{PGCN2019ICCV} and MUSES\cite{Liu_2021_CVPR} in our experiments with audio.

\noindent \textbf{Optimization:} We train all the architectures except PGCN and GTAD with their original setting. We use a batch size of 256 for PGCN and 16 for GTAD. For training, we use 4 GeForce 2080Ti 11GB GPUs. The entire codebase is based on Pytorch library except for VGGish~\cite{hershey2017cnn} which is based on Keras.

\section{\uppercase{Experiments}}

\subsection{Datasets}
\label{sec:datasets}

To compare our results with the SOTA architectures in which we incorporate audio, we evaluate our models on two benchmark datasets for temporal action localization. 

\textbf{Thumos14}~\cite{THUMOS14} contains $1010$ untrimmed videos for validation and $1574$ for testing. Of these, $200$ validation and $213$ testing videos contain temporal annotations spanning $20$ activity categories. Following the standard setup~\cite{Xu_2020_CVPR,PGCN2019ICCV}, we use the $200$ validation videos for training and the $213$ testing videos for evaluation. 

\textbf{ActivityNet-1.3}~\cite{Heilbron_2015_CVPR} contains 20k untrimmed videos with $200$ action classes between its training, validation and testing sets. Once again, following the standard setup~\cite{Xu_2020_CVPR,PGCN2019ICCV} , we train on $10024$ videos and test on the $4926$ videos from the validation set. 

\subsection{Evaluation Protocol}
\label{sec:evalprotocol}

We label a temporal (proposal) prediction (with associated start and end time) as correct if (i) its Intersection Over Union (IOU) with ground-truth exceeds a pre-determined threshold (ii) proposal's label matches the ground truth counterpart. Following standard protocols, we evaluate the mean Average Precision (mAP) scores at IOU thresholds from $0.3$ to $0.7$ with a step of $0.1$ for THUMOS14~\cite{THUMOS14} and $\{0.5,0.75,0.95\}$ for ActivityNet-1.3~\cite{Heilbron_2015_CVPR}.

\section{\uppercase{ Results}}
\label{sec:results}

\begin{table}[!t]
\centering
\resizebox{\linewidth}{!}
{
\centering
\begin{tabular}{c|c|c|c|c|c|c|c}
       & &  & \multicolumn{5}{c}{mAP@IoU} \\
\toprule
 Fusion Type(scheme) & Visual  & Audio & $0.3$ & $0.4$ & $\mathbf{0.5}$ & $0.6$ & $0.7$ \\
\toprule
 Video-only & I3D & -- & $67.91$ & $63.15$ & $56.16$ & $46.07$ & $29.89$ \\ 
  \midrule
Audio-only & -- & VGGish & 8.23 & 6.61 & 4.73 & 3.05 & 1.30 \\
\toprule
 Concat (Encoding) & I3D & VGGish & $\textbf{70.18}$ & $\textbf{64.98}$ & $\textbf{57.18}$ & $45.42$ & $28.86$  \\
 \midrule
 Dup-Trim (Encoding) & I3D & VGGish & 69.25 & 64.22 & 56.53 & \textbf{46.08} & \textbf{30.73} \\
  \midrule
 Avg-Trim (Encoding) & I3D & VGGish & 65.10 & 60.47 & 53.92 & 42.87 & 28.09 \\
  \midrule
 RM-Attention (Encoding) & I3D & VGGish & 67.88 & 63.05 & 56.19 & 45.27 & 29.98 \\
  \midrule
 Proposal (Decision) & I3D & VGGish & 52.22	& 47.42	& 39.37	& 30.54	& 17.36   \\

 \bottomrule
\end{tabular}
}
\caption{[THUMOS14] mAP for MUSES\cite{Liu_2021_CVPR} + I3D\cite{8099985} architecture of video-only, audio only and audio-visual fusional approaches.}
\label{tab:topav-thumos}
\end{table}

The mAP@0.5 results of the best fusion approach for each video-only baseline can be seen in Table~\ref{tab:archsetup}. It can be seen that incorporating audio consistently improves performance across all approaches. In particular, this incorporation results in a new state-of-the-art result on both the benchmark TAL datasets. In terms of fusion approaches, the clear dominance of encoding fusion scheme (Section~\ref{sec:encfusion}) can be seen. The Residual Multimodal Attention mechanism from this scheme enables best performance for the relatively larger ActivityNet-1.3 dataset. Similarly, our mechanism of resampling the audio modality followed by concatenation of per-modality features enables best performance for the THUMOS14 dataset. A fuller comparison of the existing best video-only and best audio-visual results obtained via our work can be seen in  Tables~\ref{tab:topav-thumos},\ref{tab:topav-anet}. The results once again reinforce the utility of audio for TAL.

\subsection{Ablations}

To analyse the utility of the fusion schemes proposed in Section~\ref{sec:fusionschemes}, we compared their performances with audio-only  and video-only methods for the best performing approach in each dataset. Looking at Tables~\ref{tab:topav-thumos} and \ref{tab:topav-anet}, it is readily evident that fusing audio and video gives the best results. Specifically, RM-Attention fusion enables best result for ActivityNet-1.3 while simple concat works best for Thumos14. The reason for simple concat's superior performance for Thumos14 can be explained by the fact that audio content is less informative regarding the action boundaries in Thumos14 compared to ActivityNet-1.3. This is also evident from the audio-only baselines -- compare the second rows of Tables~\ref{tab:topav-thumos},\ref{tab:topav-anet}. We hypothesize that RM-Attention is more effective at fusing the modalities than filtering out noise when audio modality is uninformative.  In contrast, for simple concat, the separate, non modulated contribution of audio and visual features makes the fusion scheme less susceptible to noise in any one modality. 

\begin{table}[!t]
\centering
\resizebox{\linewidth}{!}
{
\centering
\begin{tabular}{c|c|c|c|c|c|c}
       & &  & \multicolumn{4}{c}{mAP@IoU} \\
\toprule

 Fusion Type (scheme) & Visual  & Audio & $\mathbf{0.5}$ & $0.75$ & $0.95$ & avg. \\
\toprule
  Video-only & TSP & -- & $51.26$ & $37.12$ & $9.29$ & $35.01$ \\ 
  \midrule
  Audio-only & -- & VGGish & 43.07 & 28.27 & 5.82 &  28.19 \\
   \midrule
   Concat (Encoding) & TSP & VGGish & 52.6 & 37.55 & 9.19 & 36.37 \\
   \midrule
   Dup-Trim (Encoding) & TSP & VGGish & 52.31 & 37.66 & 9.49 & 36.47 \\
   \midrule
   Avg-Trim (Encoding) & TSP & VGGish & 51.91 & 37.53 & \textbf{9.55} & 36.33 \\
   \midrule
  RM-Attention (Encoding) & TSP & VGGish & $\textbf{54.34}$ & $\textbf{37.66}$ & $9.29$ & $\textbf{36.82}$  \\
 \midrule
   Proposal (Decision) & TSP & VGGish &  51.6	& 37.34	& 9.41 & 35.95   \\
 \bottomrule
\end{tabular}
}
\caption{[ActivityNet-1.3] mAP for GTAD\cite{Xu_2020_CVPR} + TSP\cite{DBLP:journals/corr/abs-2011-11479} of video-only, audio-only and audio-visual fusional approaches.}
\label{tab:topav-anet}
\end{table}

DupTrim seems to perform better than AvgTrim, while both are inferior to simple concatenation. This indicates that preserving the ideal frame rate for each modality may not be that crucial to performance and it is probably better to extract features at the same rate for each modality rather than artificially making them equal after extraction. Among the fusion schemes, proposal fusion performs the worst for both ActivityNet-1.3 and Thumos14. This is to be expected considering the fact that it just selects the best proposal out from the audio and visual streams. 

The performances of the audio-only baselines for each dataset suggests that audio information present in ActivityNet-1.3 is much more indicative of activity boundaries compared to that in Thumos14. This is also consistent with the degree of improvement due to fusion for both datasets. 

\subsection{Class-wise analysis} 

\begin{figure*}[!t]
\centering
\includegraphics[width=\textwidth]{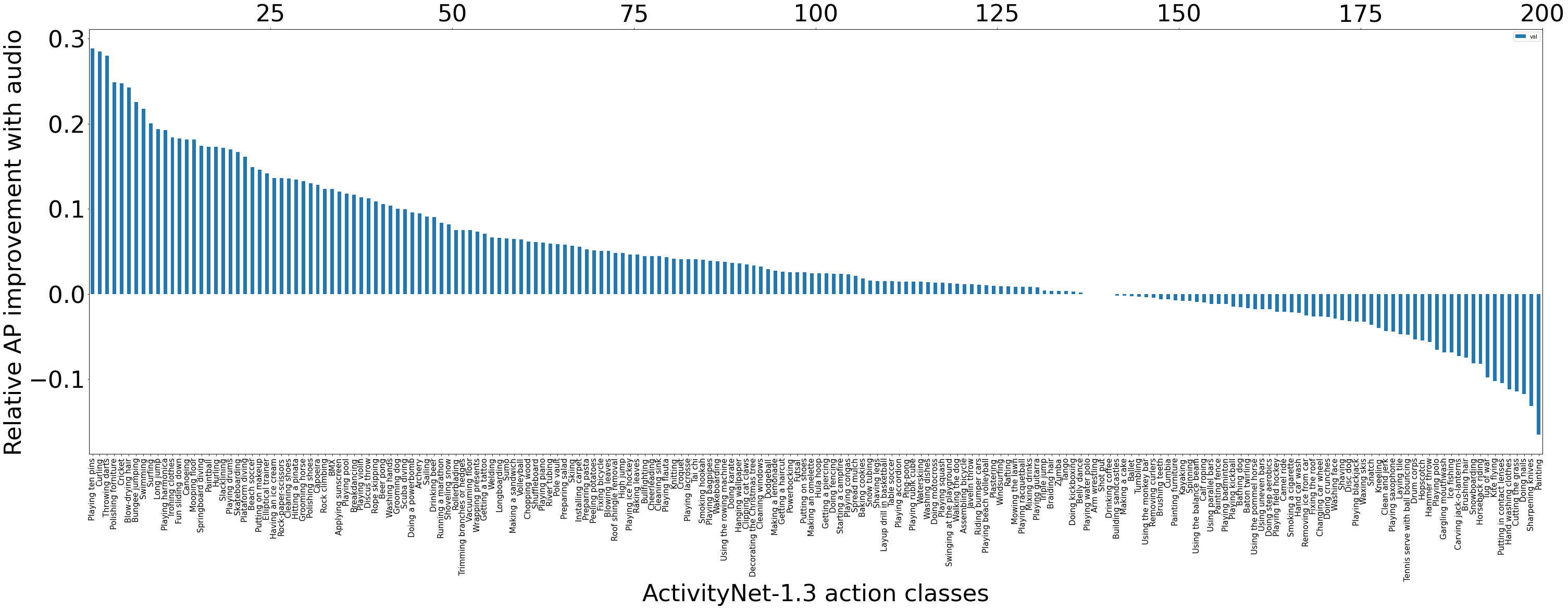}
\caption{[ActivityNet-1.3] Relative change in per-class AP of the best multimodal setup (Table~\ref{tab:archsetup}) with inclusion of audio.} 
\label{fig:avnetcw}
\end{figure*}

To examine the effect of audio at the action class level, we plot the change in Average Precision (AP) relative to the video-only score for the best performing setup. Figure~\ref{fig:avnetcw} depicts the plot for ActivityNet-1.3, sorted in decreasing order of AP improvement. The majority of action classes show positive AP improvement. This correlates with observations made in the context of aggregate performance (Table~\ref{tab:archsetup}, Table~\ref{tab:topav-anet}).
The action classes which benefit the most from audio (e.g. `Playing Ten Pins', `Curling', `Blow-drying hair') tend to have signature audio transitions marking the beginning and end of the action. The classes at the opposite end (e.g. `Painting', `Doing nails', `Putting in conact lenses') are characterized by very little association between visual and audio modalities. For these classes, we empirically observed that ambient background sounds are present which induce noisy features. However, the gating mechanism enabled by Residual Multimodal Attention ensures that the effect of such noise from the modalities is appropriately mitigated. This can be seen from the smaller magnitude of drop in AP.

\begin{figure*}[!t]
\centering
\includegraphics[width=\textwidth]{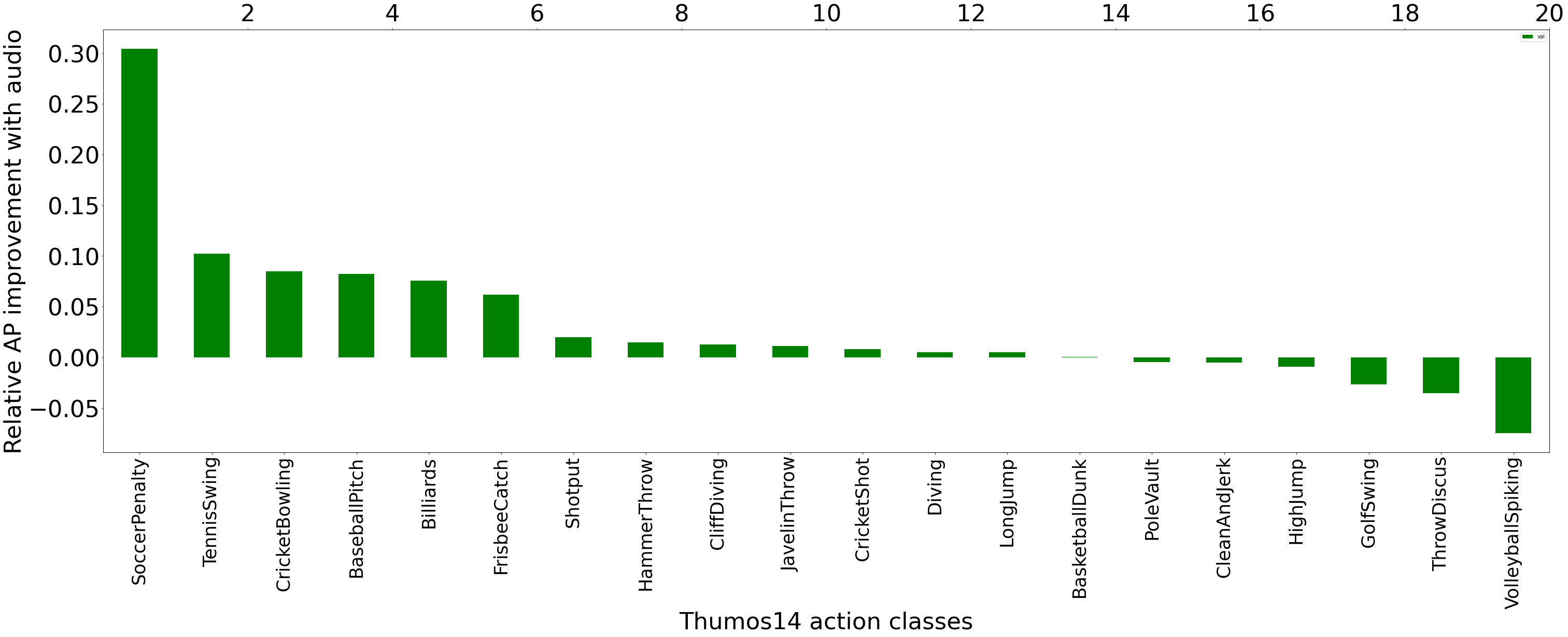}
\caption{[THUMOS14] Relative change in per-class AP of the best multimodal setup (Table~\ref{tab:archsetup}) with inclusion of audio.} 
\label{fig:thumoscw}
\end{figure*}

Figure~\ref{fig:thumoscw} depicts the sorted AP improvement plot for the relatively smaller Thumos14 dataset. Similar qualitative trends as for ActivityNet-1.3 mentioned earlier can be seen, i.e. signature audio transitions characterizing largest AP improvement classes and weak inter-modality associations characterizing least AP improvement classes. However, as mentioned in previous section, the relatively weak association between audio and video modalities in Thumos14 causes the \% of categories which are negatively impacted by audio inclusion to be greater compared to ActivityNet-1.3.

\subsection{Instance-wise analysis} 

\begin{figure*}[!t]
\centering
\includegraphics[width=\textwidth]{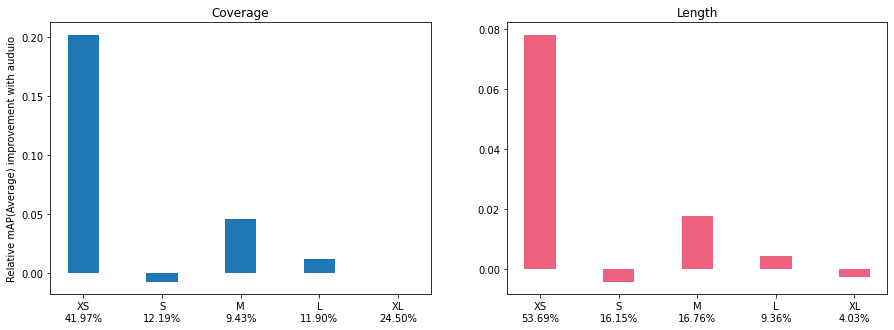}
\caption{[ActivityNet-1.3] Relative change in average mAP of the best multimodal setup (Table~\ref{tab:archsetup}) classified by instance length and coverage, with inclusion of audio. The numbers below X-labels represent the percentage of each type of instance class in the dataset} 
\label{fig:avnetit}
\end{figure*}

\begin{figure*}[!t]
\centering
\includegraphics[width=\textwidth]{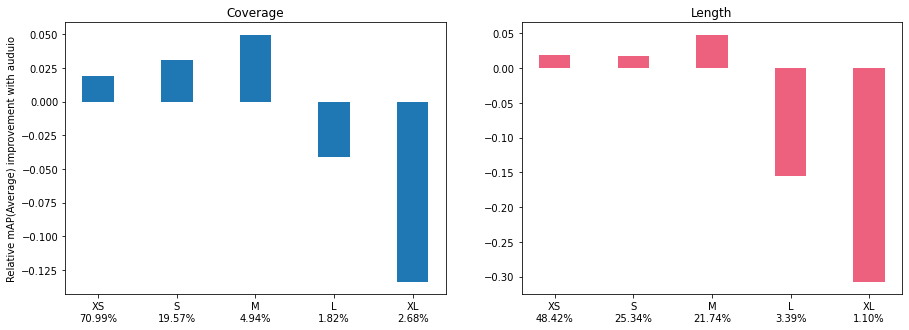}
\caption{[THUMOS14] Relative change in average mAP of the best multimodal setup (Table~\ref{tab:archsetup}) classified by instance length and coverage, with inclusion of audio. The numbers below X-labels represent the percentage of each type of instance class in the dataset} 
\label{fig:thumosit}
\end{figure*}

\begin{figure*}[!t]
\centering
\includegraphics[width=\textwidth, height = 7.2cm]{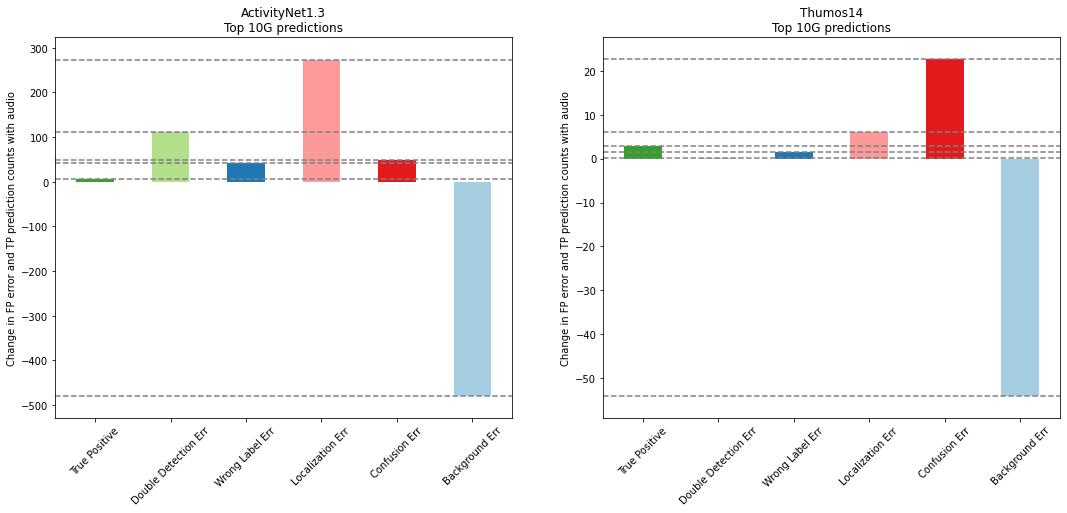}
\caption{Change in number of True Positive (TP) predictions and False Positive (FP) errors of each type of the best multimodal setup (Table~\ref{tab:archsetup}) for each dataset with the inclusion of audio. The dashed lines are added to distinguish vey close values.} 
\label{fig:fpat}
\end{figure*}

Modifying the approach used by Alwassel et al. ~\cite{alwassel_2018_detad} for their approach (DETAD), we analyze two salient attributes of data to analyse the effect of adding audio. These attributes are (i) coverage - the proportion of untrimmed video that the ground truth instance spans (ii) length (temporal duration)

To measure coverage, we normalize duration of an action instance relative to the duration of the video. Thus, larger the coverage, larger the extent the instance occupies in the video. Note that coverage lies between $0$ and $1$. We group the resulting coverage values into five buckets: Extra Small (XS: (0, 0.2]), Small (S: (0.2, 0.4]), Medium (M: (0.4, 0.6]), Large (L:(0.6, 0.8]), and Extra Large (XL: (0.8, 1.0]). The length is measured as the instance duration in seconds. We create five different length groups: Extra Small (XS: (0, 30]), Small (S: (30, 60]), Medium (M: (60, 120]), Long (L: (120, 180]), and Extra Long (XL: $>$ 180).

From the numbers below the bucket labels on x-axis in  Figures~\ref{fig:avnetit} and ~\ref{fig:thumosit}, we see that most action instances fall in Extra Small buckets. Also, the distributions of coverage and length of the ground truth instances are skewed towards the left (shorter extents).

The change in mAP due to inclusion of audio can be viewed in Figures~\ref{fig:avnetit} and ~\ref{fig:thumosit} on a per-bucket basis. The overall gain in performance for both dataset is well explained by the overwhelmingly large proportion of the total instances showing improvement due to audio : $63.3\%$ by coverage and $79.81\%$ by length for ActivityNet-1.3 and $95.5$\% by coverage and length for Thumos14. 

From the figures, we see that audio fusion enables consistent improvements for XS and M instances for both datasets while for XL instances, the mAP decreases or remained unchanged. This can be attributed in part to the fact that the shorter instances have an audio `signature' for the action that spans the majority of the instance which assists detection. For the longer action instances, the action-characteristic audio spans a small section of the instance which might not aid detection as much.

\subsection{False Positive analysis} 

Following~\cite{alwassel_2018_detad}, we consider the following error sub-categories within false positive predictions:
\begin{itemize}
    \item Double Detection error: $IoU > \alpha$ and correct label but not the highest IoU.
    \item localization error: $0.1 < IoU < \alpha$, correct label
    \item Confusion error: $0.1 < IoU < \alpha$, wrong label
    \item Wrong Label error: $IoU > \alpha$, wrong label
    \item Background error: $IoU < 0.1$ 
\end{itemize}

The change in distribution of possible prediction outcomes with inclusion of audio can be viewed in Figure~\ref{fig:fpat}. The false positive errors except Background error have increased. However, their relative frequency is smaller. The large decrease in number of Background errors more than mitigates the combined increase in other error sub-categories, explaining the overall improvement in performance. The trends in false positive errors also suggest that audio information is most useful in discriminating between activity instances and the background in untrimmed videos. In addition, we observe that the number of true positive predictions  (prediction with highest $IoU$, with $IoU > \alpha$ and correctly predicted label, where $\alpha$ is the $IoU$ threshold) increase for both THUMOS14 and ActivityNet-1.3, with the inclusion of audio.

\section{Conclusion}

In this paper, we have presented multiple simple but effective fusion schemes for incorporating audio into existing video-only TAL approaches. To the best of our knowledge, our multimodal effort is the first of its kind for fully supervised TAL. An advantage of our schemes is that they can be readily incorporated into a variety of video-only TAL architectures -- a capability we expect to be available for future approaches as well. Experimental results on two large-scale benchmark datasets demonstrate consistent gains due to our fusion approach over video-only methods and state-of-the-art performance. Our analysis also sheds light on the impact of audio availability on overall as well as per-class performance. Going ahead, we plan to expand and improve the proposed families of fusion schemes.

{\small
\bibliographystyle{ieee_fullname}
\bibliography{egbib}
}

\end{document}